\documentclass[conference]{IEEEtran}
\IEEEoverridecommandlockouts
\usepackage{cite}
\usepackage{amsmath,amssymb,amsfonts}
\usepackage{algorithmic}
\usepackage{graphicx}
\usepackage{textcomp}
\usepackage{makecell}
\usepackage{xcolor}
\def\BibTeX{{\rm B\kern-.05em{\sc i\kern-.025em b}\kern-.08em
    T\kern-.1667em\lower.7ex\hbox{E}\kern-.125emX}}

\usepackage[colorinlistoftodos]{todonotes}

\usepackage{enumitem}
\usepackage{adjustbox}
\usepackage{url}

\usepackage{enumitem}
\usepackage{graphicx}

\usepackage{filecontents}
\usepackage{listings}
\usepackage{xcolor}

\definecolor{codegreen}{rgb}{0,0.6,0}
\definecolor{codegray}{rgb}{0.5,0.5,0.5}
\definecolor{codepurple}{rgb}{0.58,0,0.82}
\definecolor{backcolour}{rgb}{0.95,0.95,0.92}

\lstdefinestyle{mystyle}{
    backgroundcolor=\color{backcolour},
    commentstyle=\color{codegreen},
    keywordstyle=\color{magenta},
    numberstyle=\tiny\color{codegray},
    stringstyle=\color{codepurple},
    basicstyle=\tiny,
    breakatwhitespace=false,
    breaklines=true,
    captionpos=b,
    keepspaces=true,
    numbers=left,
    numbersep=5pt,
    showspaces=false,
    showstringspaces=false,
    showtabs=false,
    tabsize=2
}

\begin{document}

%%
%% The "title" command has an optional parameter,
%% allowing the author to define a "short title" to be used in page headers.
\title{Pretraining Billion-scale Geospatial \\ Foundational Models on Frontier}

\author{
    \IEEEauthorblockN{ Aristeidis Tsaris
    \textsuperscript{*}
    }
    \IEEEauthorblockA{\textit{\small National Center for Computational Sciences} \\
    Oak Ridge National Laboratory\\
    Oak Ridge, TN, USA
}
\and
\IEEEauthorblockN{Philipe Ambrozio Dias}
    \IEEEauthorblockA{\textit{\small Geospatial Science and Human Security} \\
    Oak Ridge National Laboratory\\
    Oak Ridge, TN, USA
}
\and
\IEEEauthorblockN{ Abhishek Potnis}
    \IEEEauthorblockA{\textit{\small Geospatial Science and Human Security} \\
    Oak Ridge National Laboratory\\
    Oak Ridge, TN, USA
}
\and
\IEEEauthorblockN{Junqi Yin}
    \IEEEauthorblockA{\textit{\small National Center for Computational Sciences} \\
    Oak Ridge National Laboratory\\
    Oak Ridge, TN, USA
}
\and
\IEEEauthorblockN{Feiyi Wang}
    \IEEEauthorblockA{\textit{\small National Center for Computational Sciences} \\
    Oak Ridge National Laboratory\\
    Oak Ridge, TN, USA
}
\and
\IEEEauthorblockN{Dalton Lunga}
    \IEEEauthorblockA{\textit{\small Geospatial Science and Human Security} \\
    Oak Ridge National Laboratory\\
    Oak Ridge, TN, USA
}
}

\maketitle

\begingroup\renewcommand\thefootnote{*}
\footnotetext{Corresponding author: tsarisa@ornl.gov}
\endgroup

\begin{abstract}

As AI workloads increase in scope, generalization capability becomes challenging for small task-specific models and their demand for large amounts of labeled training samples increases. On the contrary, Foundation Models (FMs) are trained with internet-scale unlabeled data via self-supervised learning and have been shown to adapt to various tasks with minimal fine-tuning. Although large FMs have demonstrated significant impact in natural language processing and computer vision, efforts toward FMs for geospatial applications have been restricted to smaller size models, as pretraining larger models requires very large computing resources equipped with state-of-the-art hardware accelerators. Current satellite constellations collect 100+TBs of data a day, resulting in images that are billions of pixels and multimodal in nature. Such geospatial data poses unique challenges opening up new opportunities to develop FMs. We investigate billion scale FMs and HPC training profiles for geospatial applications by pretraining on publicly available data. We studied from end-to-end the performance and impact in the solution by scaling the model size. Our larger 3B parameter size model achieves up to 30\% improvement in top1 scene classification accuracy when comparing a 100M parameter model. Moreover, we detail performance experiments on the Frontier supercomputer, America's first exascale system, where we study different model and data parallel approaches using PyTorch's Fully Sharded Data Parallel library. Specifically, we study variants of the Vision Transformer architecture (ViT), conducting performance analysis for ViT models with size up to 15B parameters. By discussing throughput and performance bottlenecks under different parallelism configurations, we offer insights on how to leverage such leadership-class HPC resources when developing large models for geospatial imagery applications.

\end{abstract}


\section{Introduction}

% - Geospatial/RS intro: application, data types, challenges
Automated analysis of Earth Observation (EO) data using artificial intelligence (AI) tools is emerging as one of the cornerstones enabling fast and cheap exploitation of geospatial information to describe and assess physical features as well as geographically referenced events on Earth. EO data entails capturing information about the Earth's surface using sensors mounted on e.g. satellites and in-situ instruments. Deep neural network (DNN) models have significantly advanced the state of the art in automated analysis of remote sensing (RS) data, enabling extensive applications in human dynamics, precision agriculture, disaster management, humanitarian assistance, and national security. 

% Unlike traditional natural images used in CV benchmarks, EO data presents unique characteristics and challenges that include spatial & temporal awareness, data volumes & diversity, and multimodal reasoning. Importantly, appropriately addressing these challenges hold the potential for groundbreaking applications benefiting human and environmental well-being.

% Current RS constellations are mapping the Earth’s surface at paces that lead to more than 100 trillion pixels a day, with single satellites generating more than 100 terabytes (TB) a day. While this sheer volume and diversity of RS archives represent a great potential for a wide variety of applications and the development of large-scale models, 

New sensing modalities with complementary imaging characteristics are becoming operational, monitoring larger extents of the Earth's surface with improved temporal cadence and finer spatial resolution. While this presents great potential for a wide variety of applications, the sheer volume and diversity of RS archives continues to stretch the limits of human analysts and existing AI tools. Similar to other AI application domains, deep learning models currently employed for EO data analysis are limited in that: (i) they are task-specific, with limited generalization to unseen (out of distribution) data; (ii) they heavily rely on large volumes of manually annotated data samples for model training, which implies on high costs for model development.

Foundation models (FMs) are a recent breakthrough with demonstrated potential to address such limitations \cite{bommasaniOpportunitiesAndRisks}. FMs can be defined as large models (ranging from $10^8$ to $10^{12}$ of parameters) usually trained through self-supervised learning (SSL) on large volumes of unlabeled data, such that they learn generalizable features that can then be adapted at lower-costs for a wide variety of downstream tasks. First demonstrated for natural language processing (NLP), FMs are extending their impacts in other domains such as computer vision, across a variety of data types. 
A common practice is to fine-tune pre-existing FMs to align with particular use cases, but in the context of geospatial tasks, the multifaceted characteristics of the data pose additional constraints. While some recent works on RS data analysis have exploited the concepts of SSL for model pretraining, such efforts have been restricted to smaller model sizes than the proportions characteristic of FMs \cite{cong2022satmae, sun2022ringmo, wang2022rvsa}.

Recent advancements towards developing state-of-art foundation models for computer vision and natural language processing tasks, have highlighted the need for substantial compute and memory for the research and development of large-scale foundation models.
 Florence\cite{yuan_florence_2021}, an FM for computer vision with 893 million parameters, was trained for 10 days on 512 NVIDIA A100 (40GB) GPUs. Contrastive Language-Image Pre-Training (CLIP)\cite{RadfordCLIP} required 12 days on 256 NVIDIA V100 GPUs for training its largest ViT-based configuration. The ALIGN: A Large-scale ImaGe and Noisy-text embedding\cite{JiaALIGN} model was trained on 1024 Cloud TPUv3 cores, while GATO\cite{reed2022generalist} with its 1.2 billion parameters was trained on 16x16 Cloud TPUv3 cores. 
%  The availability of high performance computing (HPC) resources and expertise on how to properly configure AI training workflows on such computing power are key major barriers limiting progress on the development of billion-scale FMs for geospatial data.
% In this paper, we describe a detailed performance analysis, and preliminary evaluation for pretraining FMs on geospatial data, that are billions of parameters large. Our main contributions are:

% Such demand for vast amounts of data and high performance computing (HPC) resources are key major barriers limiting progress on the development of FMs for geospatial data. Critically, beyond just accessibility to computing resources, an additional need is developing expertise and guidelines on how to effectively pair them with data and model parallel approaches to successfully enable training of large models. This underscores the importance of developing training methodologies that can be harnessed by communities aiming to control the model, enhance its performance, and tailor it to address specific application needs.

Such demand for vast amounts of data and high performance computing (HPC) resources are key major barriers limiting progress on the development of FMs for geospatial data. Critically, beyond accessibility to computing resources, expertise and guidelines on how to effectively train FMs using data and model parallel approaches remains limited to few established organizations. This underscores the importance of developing training methodologies that can be harnessed by communities aiming to control the model, enhance its performance, and tailor it to address specific application needs. 

Within this manuscript, we present performance assessments and initial appraisal of pretraining billion-scale FMs on geospatial data sets. Our primary contributions are as follows:

\begin{itemize}
    \item We present a practical guide for training billion parameter size ViT models on HPC systems, using the newest Pytorch's FSDP model sharding library. We provide image-per-second baselines for various size ViT models, and discuss compute and communication costs to consider in training FMs for geospatial application workloads.
    \item We share best practices and study bottlenecks on distributing ViT training on the Frontier HPC system while scaling to various size models. ViT has been the backbone of many vision and multi-modal foundational models.
    \item We demonstrate the benefits of training billion-scale models for geospatial data, with linear-probing experiments highlighting gains up to $+30\%$ on remote sensing imagery classification tasks across three independent datasets. This is the largest size model trained on geospatial applications. 

\end{itemize}

\section{Background and Related Work}

\noindent\textbf{Model architectures.} In contrast to hand-crafted features based methods, convolutional neural networks (CNNs) trained over large amounts of labeled data have enabled significant progress in computer vision as they learn to extract hierarchical features that better capture semantically meaningful patterns. The UNet \cite{ronneberger2015unet} encoder-decoder architecture is a representative example as it has been widely adapted for pixel-level semantic segmentation tasks of remote sensing image applications. However, the local nature of the convolution operator limits its ability to capture long-range interaction in scenarios that require larger context.
% While stacking multiple convolution layers has enabled encoder-decoder architectures (e.g., UNet []) to be successful including in pixel-level semantic segmentation tasks, the local nature of the convolution operator limits its ability to capture long-range interaction in scenarios that require larger context. 

Attention mechanisms have been explored to address such issues, with the notion of \textit{self-attention} at the core of novel architectures such as the Transformer architecture \cite{vaswani2017transformers} that has become the main reference architecture behind modern large language models (LLMs). The Vision Transformer (ViT) \cite{dosovitskiy2020vit} first extended the concept of attention-only architectures for image processing, with following studies revealing their increased performance with scale \cite{zhai2021scalingvit, dehghani2023scaling}. ViTs have thus emerged as the core architecture developing FMs for computer vision applications. % However, also noteworthy are works such as ConvNeXt \cite{woo2023convnext}, which leverage strategies developed for ViT-based models to enable SSL training of large models based on convolutional architectures.

% Transformer-based architectures are easily scalable/parallelizable, such that most current FMs consisting of Transformer-based architectures.
% - FMs for computer vision

% - mostly based on ViTs, since arch is more easily scalable. Acknowledge papers such as InternImage using CNN-based or hybrids, but argue that we focus on ViT

% - connect to RS works using ViT + SSL
% - mention contrastive learning and MAE
% - mention the ones using MAE
% - 
% \begin{itemize}
%     \item ViTs are taking over CNNs,
%     \item why they are better especially for SS, since they can enlarge receptive field, combine a lot of data res, modalities, etc.
%     \item foundation models for vision always include ViTs
% \end{itemize}
% \begin{itemize}
%     \item Describe ViTs in details, and ones used here
%     \item Describe SS methods, and MAE and scale
%     \item Describe largest today ViTs and MAEs
%     \item Vision foundation models for remote sensing
%     %bring content from our IGARSS paper here.
% \end{itemize}

In contrast to preceding deep learning paradigms, FMs have distinct characteristics with respect to: (i) \textit{scale} and \textit{scope}, in terms of model size as well as pretraining with internet-scale data volumes; (ii) extraordinary \textit{transfer learning} capabilities with fewer labeled datasets on a %, as such models are able to extract features that can be efficiently adapted to 
a wide range of downstream tasks. 

\noindent\textbf{Pretraining mechanisms.} Self-supervised learning (SSL) at scale is a key ingredient enabling the success of FMs. By leveraging surrogate tasks as sources of supervision when no labels are available, SSL unlocks the potential of learning from unprecedentedly large unlabeled datasets for which manual labeling would be unfeasible. Two main approaches prevail for SSL of computer vision models. As exemplified by the SimCLR framework\cite{chen2020simple}, contrastive learning schemes exploit correlated views of the same image constructed through augmentations (e.g., random cropping and color distortions). During training, the model is tasked to maximize feature similarity between pairs augmented from the same image, while minimizing similarity between examples originating from different images. Analogous to the masked language model used for training NLP models \cite{devlin2019bert}, the denoising concept of masked autoencoders (MAEs) \cite{he2022masked} enforces the model to reconstruct pixels of masked image patches based on the remaining visible image patches - this approach has been demonstrated to yield models with strong generalization capabilities.

\noindent\textbf{Foundation Models for Remote Sensing.} While contrastive learning methods for natural images typically rely on artificial augmentation to construct positive pairs, in the remote sensing domain approaches can perform SSL by aligning representations extracted from different image acquisitions covering the same location but from different timestamps and/or sensors \cite{ayush2021geography}. More recently, multiple strategies targeting FMs for RS have been introduced leveraging MAE to pretrain larger Transformer-based architectures. The SatMAE described in \cite{cong2022satmae} uses MAE to pretrain a ViT-L using optical imagery from the Functional Map of the World (fMoW) dataset \cite{christie_fmow_2017}, demonstrating its applicability on land cover classification and building segmentation tasks. The concept of MAE-based pretraining of ViT for RS is likewise exploited in \cite{wang2022rvsa}, which proposes the idea of rotated varied-size attention (RVSA) to enhance the robustness of ViTs to the variations in size and orientation characteristic of objects when captured in remote sensing imagery. Similarly, the RingMo framework \cite{sun2022ringmo} adopts masked modeling at pixel-level for model pretraining, with patches that are only partially masked as a mechanism targeting better reconstruction and characterization of smaller structures.

Crucially, all such works toward FM for RS data have been restricted to models with less than $300M$ parameters. RVSA \cite{wang2022rvsa} and RingMo \cite{sun2022ringmo} perform pretraining on datasets containing millions of images, but are both restricted to models based on the ``base`` configuration of ViT that is the order of $<100M$ parameters. While the SatMAE \cite{cong2022satmae} work includes ViT-Large architectures, the approximately $300M$ parameters of that architecture are still far smaller than the $1B+$ models powering FMs for other computer vision application domains.

%Establish need for research in scalability for foundation models
In addition to the model size scaling challenge, FMs for RS involves training over large corpus of data for them to learn non-trivial features that are generalizable to a variety of downstream tasks. The RVSA exploited the MillionAID dataset \cite{long2021millionaid} for model pretraining. Introduced for image classification tasks, the MillionAID dataset contains over one million RS scenes from various sensors and resolutions, with image sizes ranging from $110\times110$ to $31k\times31k$ pixels. The fMoW dataset \cite{christie_fmow_2017} used by SatMAE \cite{cong2022satmae} contains over 500k optical image patches collected from multiple sensors, while \cite{sun2022ringmo} reports experiments using a custom (not publicly-available) dataset containing over $2M$ images.

% Given the variability in RS imagery, FMs for RS are required to be trained over large corpus of data for them to learn non-trivial features that are generalizable and can later be leveraged by a variety of downstream tasks. In addition to the challenge posed by the massive amount of training data, the models themselves are huge with millions of parameters to be learnt during the course of training. Although technological advancements have resulted in widespread use of high-performance computing systems with powerful compute capabilities and vast storage, it is not feasible to subscribe to the idea of unlimited compute and storage. Given the increasing gap between demand and supply of both memory and compute\cite{sevilla2022compute}, there arises a need to develop strategies aimed at optimizing resources for training and deploying foundation models. 

\noindent\textbf{Distributed deep learning.} It is becoming evident as RS workloads grow that the pretraining of FMs will continue to demand enormous compute power and time in addition to substantial data storage. Although technological advancements have resulted in widespread use of HPC systems with powerful compute capabilities and vast storage, it is not feasible to subscribe to the notion of unlimited compute and storage. Distributed optimization strategies are there sought after to enable the growing compute at scale needs. This paradigm seeks to accelerate model training to achieve higher throughput by leveraging efficient optimization techniques and high-performance computing. 
%and learning refers to the paradigm of accelerating the model training to achieve higher throughput by leveraging efficient optimization techniques and high-performance computing. 

Parallelization strategies have become essential when training large deep neural network models whose memory footprint exceeds the memory of a single GPU device. These strategies %under the padardigm of distributed deep learning, 
can be broadly categorized into: (i) conventional perspective that involves data and model parallelism; and (ii) modern perspective that involves intra-operator and inter-operator parallelization \cite{zheng2022alpa}. In this work we leverage the PyTorch's native Fully Sharded Data Parallel (FSDP) strategy \cite{zhao2023pytorch}, which takes inspiration from the Zero Redundancy Optimizer\cite{ren2021zero} that was first implemented in DeepSpeed \cite{10.1145/3394486.3406703} and later had an improved version implemented in PyTorch.

\noindent\textbf{Evaluation protocols for FMs.} Since the main targeted capability of FMs is the ability to extract features that are easily generalizable to multiple downstream tasks, evaluation protocols adopted by related works \cite{bommasaniOpportunitiesAndRisks, cong2022satmae,sun2022ringmo, wang2022rvsa,yuan_florence_2021} are often based on the concepts of fine-tuning and linear probing. 
By their nature, FMs are very large, such that fine-tuning fewer tasks-specific layers is preferable from the perspective of computational and labeled-data budgets. Fine-tuning configurations can range between updating all layers of the model during downstream specialization, freezing specific layers/blocks (e.g., the pretrained backbone), to the linear probing configuration where all pretrained parameters are frozen and parameter updates are thus restricted only to a final classification layer appended to the pretrained model.

Results are typically contrasted to baselines trained in fully supervised schemes for multiple tasks. Related works adapting FM models for the RS domain typically conduct the evaluation on three main tasks: image classification, object detection, and semantic segmentation.

\section{Experimental Setup} \label{sec:fsdp}
In this section we discuss the model architectures adopted for experimentation, the hardware characteristics of the Frontier supercomputer, and the different sharding strategies enabled by the PyTorch's FSDP library. 

\subsection{Model Architecture Variants} \label{sec:archs}

% As aforementioned, most FMs for computer vision adopt variants of the ViT architecture as a backbone, while some form or mask-autoencoder is present in the recently developed architectures. The MAE architecture was used for pre-training and evaluating the impact of scaling the model size. The performance analysis was studied in both the MAE and ViT architectures.

% The encoder of the MAE is basically the ViT architecture (usually refers as ViT backbone of MAE), while the additional decoder head is a stuck of transformer blocks used only in the pre-training phase. As it was described in the original MAE work \cite{he2022masked}, the MAE decoder is responsible for only a small fraction of the overall compute (e.g., $<10\%$ of FLOPS per token as compared to a ViT-L encoder), and it is used only in the pre-training phase. Also, it is common \cite{he2022masked} when the parameters of the models are increased, the decoder is kept constant, and so effectively it only the ViT backbone that is changed.  In light of this, and in order to be more general in our performance baselines, for the throughput measurements we mostly focus on the ViT variants.

 Table \ref{table:models} summarizes the different ViT variants explored in the remainder of this work. The ViT base (ViT-Base) and huge (ViT-Huge) variants follow the original vision Transformer paper \cite{dosovitskiy2021image}, with the ViT-Base containing 87M parameters, an embedding size (\textit{width}) of 768, 12 encoder layers (\textit{depth}), 12 heads per self-attention layer, configured with input patches $16\times16$ pixels large. For the ViT-Huge model and all listed billion-scale models, we adopt instead input patches that are $14\times14$ pixels large as per \cite{dosovitskiy2021image} and related works. 
 
 For the pretraining we used the MAE architecture. As discussed in the original MAE work \cite{he2022masked}, the decoder architecture for the image reconstruction pretraining task can be flexibly designed, with the MAE decoder being responsible for only a small fraction of the overall compute (e.g., $<10\%$ of FLOPS per token as compared to a ViT-L encoder). Results from \cite{he2022masked} shows that a lightweight decoder with 8 Transformer blocks and a width of 512 is sufficient for pretraining of ViT models with strong capabilities for linear probing. We similarly adopt such default configuration to define the decoder of our MAE architectures for ViT pretraining, with a loss function based on the mean squared error (MSE) between model reconstruction and original image, for normalized pixel values of each masked patch.
 
 We use the notation ViT-$x$B to denote our models with $x$ billion parameters. Scaling studies in \cite{dosovitskiy2020vit} note that it is most effective to scale all aspects (depth, width, MLP-width, and patch-size) by similar amounts, while \cite{zhai2021scalingvit} performs extensive simulations to define ViT variants of different sizes. Following those empirical guides and \cite{dehghani2023scaling}, which has trained the largest ViT available today (22B parameters), we increase the number of encoder layers and heads by gradually scaling the embedding size from 768 to 5040.
 
 \begin{table}[h!]
\centering
\caption{Vision Transformer (ViT) Model Architectures \\ Explored in this Work}
\begin{tabular}{ |c|c|c|c|c|c| }
 \hline
 \textbf{Model} & \textbf{Width}  & \textbf{Depth} & \textbf{MLP} & \textbf{Heads} & \textbf{Parameters [M]} \\
 \hline
 ViT-Base &   768  & 12  & 3072 & 12  & 87 \\
 ViT-Huge &   1280  & 32  & 5120 & 16  & 635 \\
 ViT-1B &   1536  & 32  & 6144 & 16  & 914 \\
 ViT-3B &   2816  & 32  & 11264 & 32  & 3067 \\
 ViT-5B &   1792  & 56  & 15360 & 16  & 5349 \\
 ViT-15B &   5040  & 48  & 20160 & 48  & 14720 \\
 \hline
\end{tabular}
\label{table:models}
\end{table}

\subsection{Hardware and Software}
All our experiments were performed using the Frontier Supercomputer \cite{FrontierWebsite} at the Oak Ridge Leadership Computing Facility. Each Frontier node has a single 64-core AMD EPYC CPU, and four AMD Instinct MI250X GPU accelerators. The MI250X GPU is comprised of two Graphics Compute Dies (GCDs), connected with Infinity Fabric CPU-GPU, while the four MI250X GPUs are connected with Infinity Fabric GPU-GPU of $50 GB/s$. The system identifies each GCD independently, so from the application perspective it can be considered that each node has 8 GPUs, each with 64 GB of high-bandwidth memory. For simplicity we are going to use the term GPU when referring to a GCD. The nodes are connected via a Slingshot-11 interconnect with $100 GB/s$, to a total of 9408 nodes, making it the first true exascale machine. For the software stack, we used PyTorch 2.1, ROCm v5.4.0, MIOpen v2.19.0, RCCL v2.13.4 with libfabric v1.15.2 plugin.

\subsection{Sharding Billion Scale Models}
We used PyTorch's Fully Sharded Data Parallel (FSDP) \cite{zhao2023pytorch} distributed strategy in order to scale both the batch size and the model size. On one GPU, with our current software stack, we can fit up to a 3B parameter ViT model. Still, the compute workload from a 100M parameter model to a 3B parameter is drastically different, such that even for models that can fit on a single GPU it is worth studying if distributing the compute to multiple GPUs will be a good trade off for the extra communication cost. 

Scaling the batch size is usually done by copying the model across ranks, and doing collective all-reduce operations during the backward pass for sharing the gradients. PyTorch's distributed data parallel (DDP) module has been widely used in the past for this distributed strategy. The FSDP mode equivalent to DDP is called NO\_SHARD, where parameters, gradients, and optimizer states are not sharded across ranks. Even though algorithmically the two approaches are the same, their implementations are different. Moreover, while most preexisting studies are based on NVIDIA GPUs, this work relies on AMD GPUs to handle a very diverse workload in terms of computation operations. Given such differences, in this work we study the impact of DDP vs FSDP-enabled sharding strategies for different model configurations.
% we think it is interesting to show the comparison. For a pure data-parallel approach we can only run up to 3B models, since 5B and 15B won't fit on a single GPU. 

We chose FSDP because it is the native solution of PyTorch for zero-redundancy parallelism, and as such it can be leveraged for different model architectures without depending on third party implementations. Since it benefits from PyTorch built-in support for both NVIDIA CUDA and AMD ROCm stacks, it is expected to be more system independent and less likely to change, compared to other frameworks. Compared to DeepSpeed, FSDP offers several modes in terms of sharding strategies and communication optimizations, which can benefit optimal usage of different hardware systems. 

\noindent\textbf{Sharding configurations.} In addition to the NO\_SHARD strategy, FSDP offers the FULL\_SHARD, SHARD\_GRAD\_OP and HYBRID\_SHARD. The FULL\_SHARD has the highest communication overhead, but produces the lowest memory footprint. It shards parameters, gradients and optimizer states across devices, offering the largest overlap between compute and communication. In contrast, the SHARD\_GRAD\_OP shards only gradients and optimizer states during computation, while the model parameters are sharded outside computation. In this sharded strategy, the communication is lighter than FULL\_SHARD but the memory footprint is increased. 

% The HYBRID\_SHARD offers the most flexibility, since it is a hybrid strategy between model shard and model replica. For example, assume we apply HYBRID\_SHARD on a single Frontier node (i.e. 8 GPUs total) allowing only the two closest GPUs to perform model shard in between. We will call this HYBRID\_2GPUs in the performance section below, and we also introduce the term full-scaling-group to refer to the number of GPUs allowing to model shard parameters in between. In this example we would have 8 GPUs total, and so we can form 4 full-scaling-groups, each with 2 GPUs. Between the two closest GPUs, all-gather and reduce-scatter communication would occur, while the model will be replicated four times, allowing to share the parameters with all-reduce communication. An example of HYBRID\_8GPUs on two nodes, would the model shards parameters within a node, so full-scaling-group of 8, and performs data-parallel across nodes. This mode is ideal for medium size models, that can fit on a single node, and reduce the communication overhead by performing only all-reduce across nodes. Since deep learning workflows are quite complicate optimization processes, and the performance depends a lot on the lower level software stack as well, we will show in the performance section below if this will result in real time application improvement for our runs.

The HYBRID\_SHARD offers the most flexibility, since it is a hybrid strategy between model shard and model replica. For example, we define a strategy named HYBRID\_2GPUs which applies HYBRID\_SHARD on a single Frontier node (i.e., 8 GPUs total) such that only the two closest GPUs perform model sharding. Let the term \textit{sharding-group} refer to the number of GPUs across which model sharding occurs. For a single node with 8 GPUs, such HYBRID\_2GPUs strategy implies on 4 sharding-groups, each with 2 GPUs. Between the two closest GPUs, all-gather and reduce-scatter communication would occur, while the model will be replicated four times with all-reduce communication enabled for sharing parameters. 

Analogously, we define a HYBRID\_8GPUs strategy that allows model sharding across all 8 GPUs of a single node. For example, using HYBRID\_8GPUs on two nodes allows forming 2 sharding-groups, with data-parallel performed across nodes. This mode is ideal for medium size models that can fit on a single node, reducing the communication overhead by performing only all-reduce across nodes. 
% Since deep learning workflows are quite complicate optimization processes, and the performance depends a lot on the lower level software stack as well, we will show in the performance section below if this will result in real time application improvement for our runs.

\section{Performance Evaluation: Scaling Model Size}\label{sec:scaling}
In this section, we discuss the measured computational costs as we scale the ViT variants. Specifically, we report memory footprint as well as throughput by measuring images-per-second ($ips$) for the six ViT models described in Table \ref{table:models}, while pursuing different FSDP-sharding strategies.

\subsection{MAE Workload Bottlenecks}
\begin{figure}[h!]
  \includegraphics[width=\linewidth]{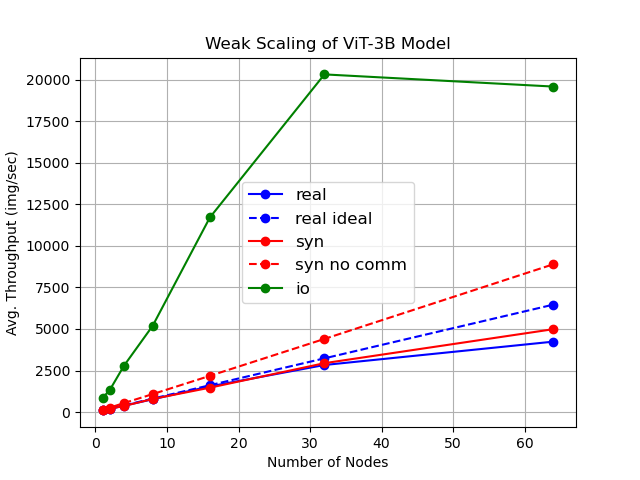}  
\caption{Weak scaling plot of the MAE 3B parameter model. The solid blue line shows the measured average image-per-second (ips) of the real application, while the dash blue line (ideal) shows the ips drawn from an ideal linear scaling scenario without any additional costs as we scale. The "syn" plot, are runs on synthetic data, represent the compute and communication performance, while the "syn no comm" is synthetic on runs as well but without any communication. The "IO" plot shows the IO performance of the application. All hyper-parameters are kept constants for the tests, with a local batch-size 32 with "NO\_SHARD" FSDP strategy.}
\label{fig:perf-io}
\end{figure}

We start by scaling our models up to the ViT-3B size for MAE-based pretraining configurations, which is the largest MAE model we can fit with our current software stack on a single Frontier's GPU. Figure \ref{fig:perf-io} shows the pretraining workload with the ViT-3B backbone for \textit{IO}, synthetic (\textit{syn}), and real application time (\textit{real}). IO corresponds to the case where the PyTorch dataloader is ran in isolation, while for the synthetic configuration we run the model on cached data. The FSDP NO\_SHARD strategy was used to scale the batch size up to 64 nodes, with a local batch size of 32, while using four data loader workers per GPU rank. 

From the comparison between \textit{io} and \textit{syn} curves we can see that even for one node the IO is faster than synthetic, with the difference between the two getting larger as we scale to 64 nodes. Therefore, we can conclude that even if we improved the IO, we would not see any benefit on the real application time, as it is compute and/or communication bound.

The \textit{syn no comm} curve refers to synthetic runs without any communication from FSDP, with its comparison of throughput against the \textit{syn} configuration providing insights on the communication costs of the application. We see that the communication cost does increase as we scale, reaching around $22\%$ for the 64 nodes runs. That indicates that while the application is closer to compute bound for smaller scale runs, communication costs take over as we increase the number of nodes.

Since it was shown the MAE workload is mostly compute and communication bound, for the rest of Section \ref{sec:scaling} the performance runs are done on synthetic data as it will provide us with more insights of the performance. Moreover, as it was discussed in \ref{sec:archs}, the ViT part of the MAE workload is the most compute-demand part and is the component of interest for model scaling, so we we are going to focus on the ViT architecture for the rest of the performance analysis.

\subsection{Communication Optimizations with FSDP}
In light of the insights from the previous sections, we investigate different FSDP sharding strategies to understand the trade-offs between compute and communication cost.

In addition to sharding options, FSDP provides several options for prefetching the parameters in the backward pass. In the \textit{None} configuration, parameters for the next layer are requested after the communication calls between all FSDP ranks. In the BACKWARD\_POST configuration the parameters are requested also before the communication calls but before the current layer drops its parameters, while the BACKWARD\_PRE is even before the communication calls. Finally, FSDP also offers the \textit{limit\_all\_gathers} option to synchronize the threads to prevent too many in-flight all-gathers.

\begin{figure}[h!]
  \includegraphics[width=\linewidth]{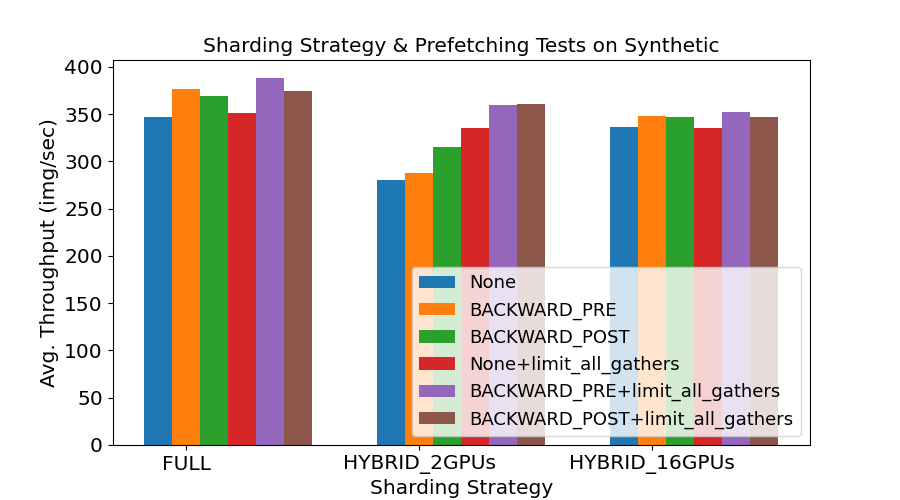}  
\caption{Average image-per-second (ips) for the ViT-5B model architecture as a function of three sharding strategies for various FSDP configurations. All hyper-parameters are kept constants for the tests, with a local batch-size 32 on 8 Frontier nodes.}
\label{fig:perf-configs}
\end{figure}

Figure \ref{fig:perf-configs} shows the performance of three FSDP sharding strategies for the ViT-5B model for various configurations on eight nodes. Here we wanted to chose a model that can't fit on a single GPU, to highlight more possible differences in the configuration. Overall, we observe that \textit{limit\_all\_gathers} option improves throughput for most configurations, with the highest gains observed for the HYBRID\_2GPUs configuration. Moreover, the BACKWARD\_PRE yields the highest throughput across most scenarios. This matches expectations since it provides the most overlap between compute and communication, although differences in performance are not very big. For the analysis that follows we fixed the parameters that provide us with the best image-per-second performance, i.e. BACKWARD\_PRE and \textit{limit\_all\_gathers}.

\subsection{Weak scaling: ViT-Base to ViT-3B}

\begin{figure*}[h!]
\centering
\begin{minipage}[t]{.33\textwidth}
  \centering
  \includegraphics[width=1.05\linewidth]{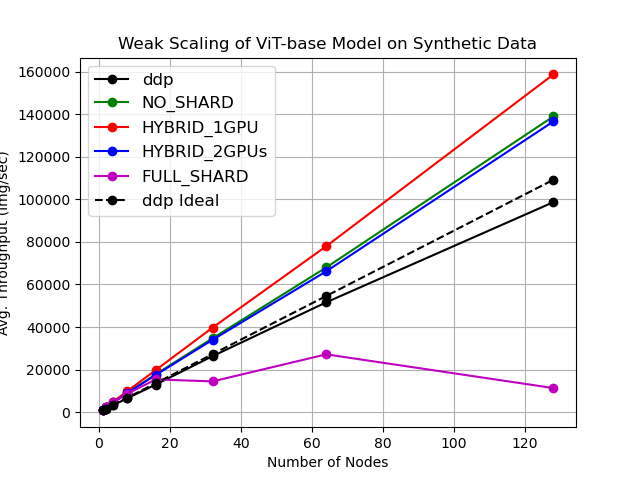}
\end{minipage} \hfill
\begin{minipage}[t]{.33\textwidth}
  \centering
  \includegraphics[width=1.05\linewidth]{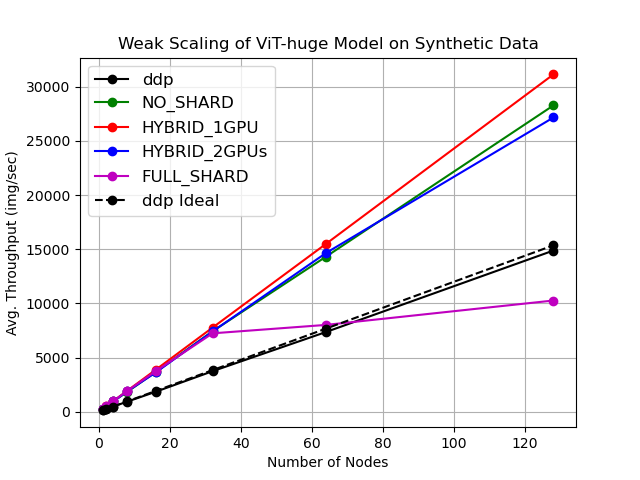}
\end{minipage}\hfill
\begin{minipage}[t]{.33\textwidth}
  \centering
  \includegraphics[width=1.05\linewidth]{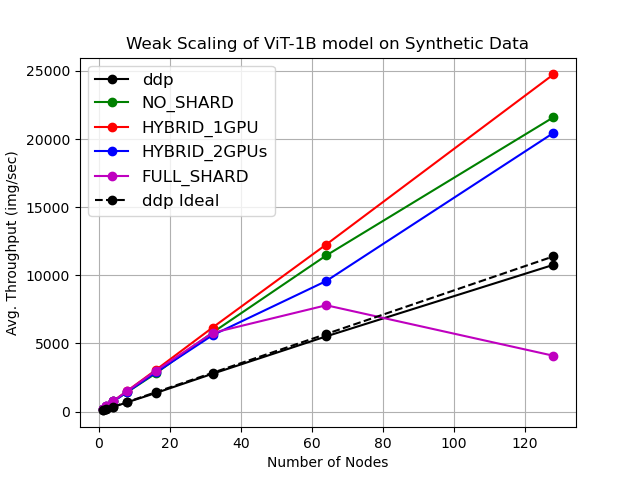}
\end{minipage}\hfill
\begin{minipage}[t]{.33\textwidth}
  \centering
  \includegraphics[width=1.05\linewidth]{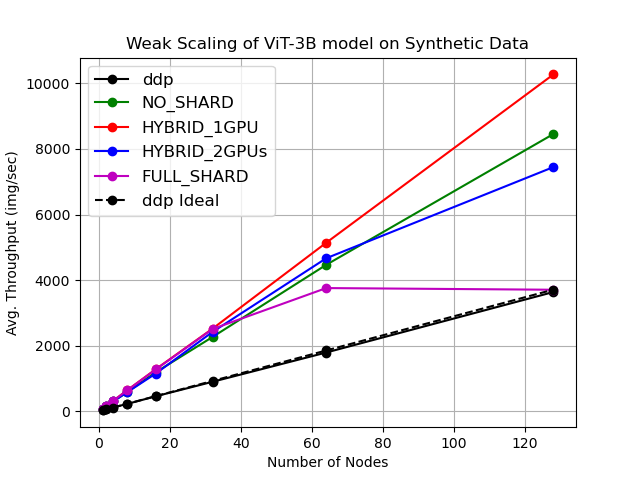}
\end{minipage}
\begin{minipage}[t]{.33\textwidth}
  \centering
  \includegraphics[width=1.05\linewidth]{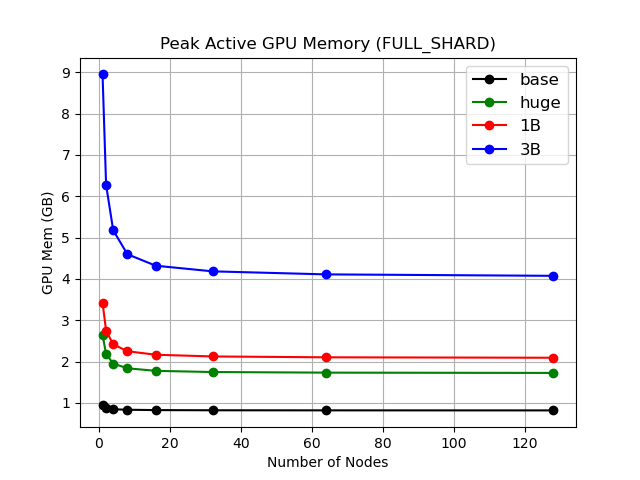}
\end{minipage}
%\begin{subfigure}{.28\textwidth}
%  \centering
%  \includegraphics[width=1.05\linewidth]{plots/perf_mem_resev_1G.png}
%\end{subfigure}
%\begin{subfigure}{.28\textwidth}
%  \centering
%  \includegraphics[width=1.05\linewidth]{plots/perf_mem_alloc_1G.png}
%\end{subfigure}
\begin{minipage}[t]{.28\textwidth}
  \centering
  \includegraphics[width=1.05\linewidth]{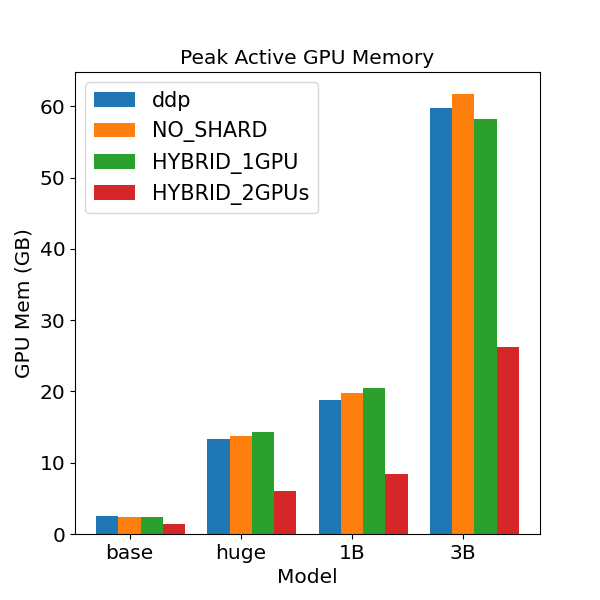}
\end{minipage}
\caption{The plots show the weak scaling for four model architectures: base (top left), huge (top center), 1B (top right) and 3B (bottom left). All four models can fit on a single GPU on Frontier. Also, the bottom center plot shows the memory usage of the FULL\_SHARD mode, and the bottom right plot for the rest of the FSDP modes. The measured average image-per-second (ips) has been measured with a local batch-size of 32 for different FSDP sharding strategies as well as the distributed-data-parallel (DDP) strategy. The dash lines (ideal) show the ips drawn from an ideal linear scaling scenario without any additional costs as we scale. The memory usage per GPU for the DDP and HYBRID is constant, as we use more nodes, while the FULL sharding strategy is not.}
\label{fig:weak_scale_1GPU}
\end{figure*}

Figure \ref{fig:weak_scale_1GPU} shows the measured averaged image-per-second (ips) for the four ViT configurations of Table \ref{table:models} that fit on a single GPU on Frontier. Here, we test the different sharding strategies explained in Section \ref{sec:fsdp} against each other as well as against the well-established data-parallel (DDP) approach. We also evaluate the HYBRID\_1GPU configuration, since although it should be equivalent with NO\_SHARD (sharding-group is 1), their implementation  PyTorch might differ and thus affect the communication and compute overlap. 

The peak memory usage is also shown in Figure \ref{fig:weak_scale_1GPU}. The HYBRID, DDP, and NO\_SHARD are constant as we use more nodes, while the FULL\_SHARD shards parameters across all ranks, and so it depends on the number of GPUs used. We see the ViT-3B model uses more than 60 GB of memory per GPU, while when the model is sharded on two GPUs, i.e. HYBRID\_2GPUs, the memory usage is dropped in half. For the FULL\_SHARD we see a much bigger a drop in memory usage, up to 4 GB for the ViT-3B model, and this is expected since the model is sharded across all the available GPUs.

The weak scaling plots in Figure \ref{fig:weak_scale_1GPU} show that the FULL\_SHARD underperforms for all model sizes at scale. As we scale the model size the FULL\_SHARD performs better as the number of nodes increases. For example, for ViT-B the performance flattens for more than 16 nodes, while for ViT-3B it only flattens after increase to 64 nodes. So we can conclude for the FULL\_SHARD strategy that when the compute is small, as we can see from the memory usage, the application becomes communication bound at a much smaller scale (as measured by number of workers) than compared to larger models with higher compute demands. 

Figure \ref{fig:weak_scale_1GPU} also shows that HYBRID\_1GPU, HYBRID\_2GPUs, and NO\_SHARD are all faster than DDP. In terms of communication footprint, the NO\_SHARD should be very similar to DDP, i.e. mostly all\_reduce communication. However, the balance between the time the communication calls are issued and their message size matters a lot as we scale different model sizes. As we keep its default parameterization, DDP does keep a constant message size for the different models, which likely becomes too small as the model size increases. Meanwhile, FSDP in general seems to be better aware of the balance between communication call time and communication size, such that as we go from the ViT-base to the ViT-3B model the gap between DDP and FSDP grows larger. It is also interesting to note the compute and communication trade-offs: FULL\_SHARD is faster at smaller scale compared to DDP, while the reverse holds as either the model size or the number of nodes increases.

Finally, another observation from Figure \ref{fig:weak_scale_1GPU} is that HYBRID\_1GPU performs better than HYBRID\_2GPUs and NO\_SHARD for all models. As shown by the memory footprint plots, the HYBRID\_1GPU uses almost twice the memory of the HYBRID\_2GPUs, which is expected since HYBRID\_2GPUs shards the model parameters across the two closest GPUs. Yet, although the compute workload should be almost half for the HYBRID\_2GPUs, the HYBRID\_1GPU performs better for all model sizes. Most likely the reason is that the overhead for synchronization between the two GPUs, along with additional communication cost for model sharding, is bigger than the benefits of less compute workload. Also, we can see from the ViT-3B the difference between the two is even larger, compared with smaller models, indicating that the application is communication bound as we scale.

\subsection{Sharding Strategies for Larger ViT Models}

\begin{figure*}[h!]
\centering
\begin{minipage}[t]{.36\textwidth}
  \centering
  \includegraphics[width=\linewidth]{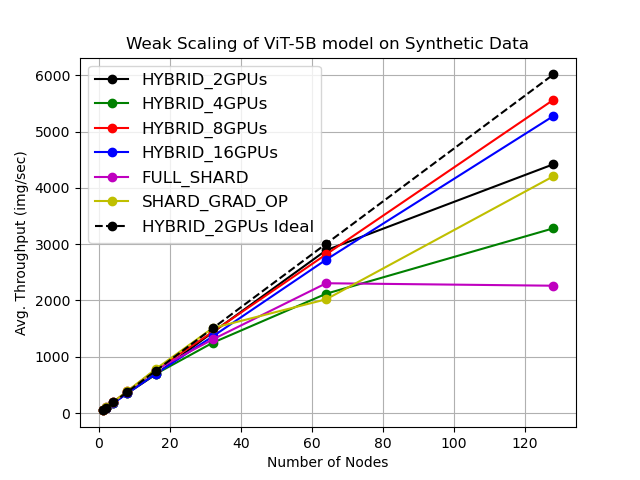}
\end{minipage}
\begin{minipage}[t]{.36\textwidth}
  \centering
  \includegraphics[width=\linewidth]{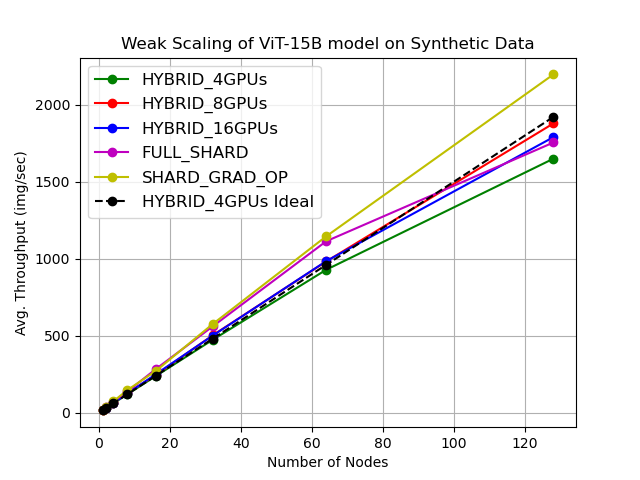}
\end{minipage}
\begin{minipage}[t]{.26\textwidth}
  \centering
  \includegraphics[width=\linewidth]{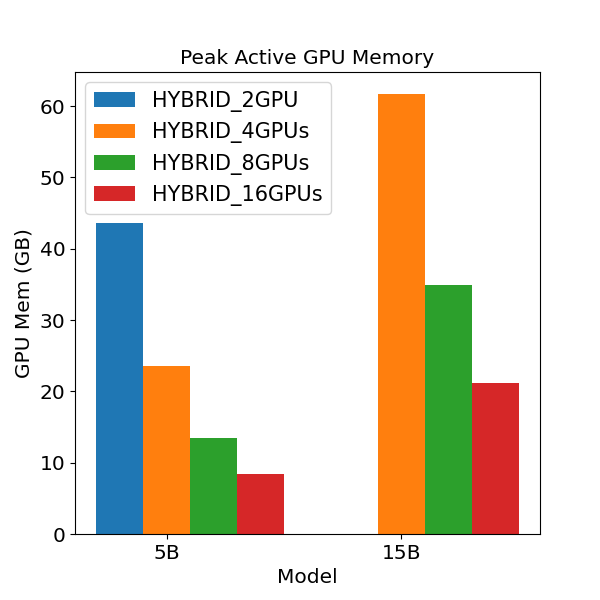}
\end{minipage}
\begin{minipage}[t]{.4\textwidth}
  \centering
  \hspace*{-.3cm}\includegraphics[width=\linewidth]{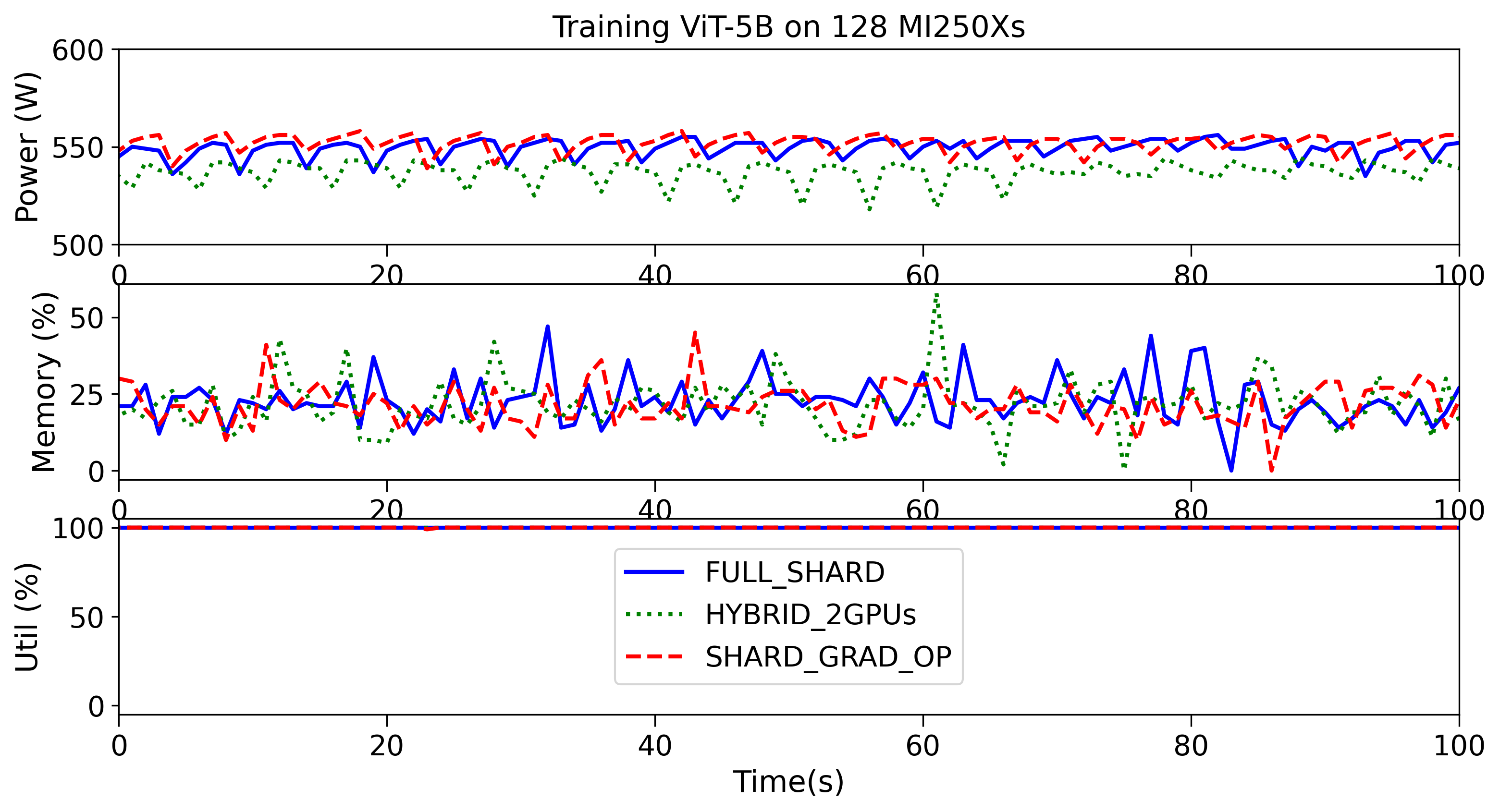}
\end{minipage}
\begin{minipage}[t]{.28\textwidth}
  \centering
  \hspace*{-.3cm}\includegraphics[width=\linewidth]{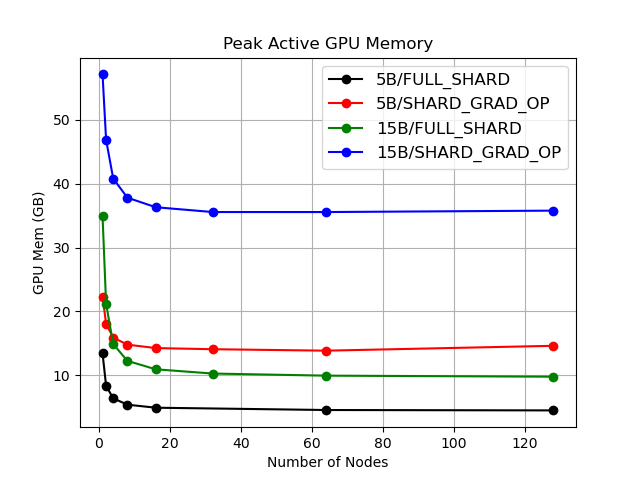}
\end{minipage}
%\begin{subfigure}{.2\textwidth}
%  \centering
%  \hspace*{-.33cm}\includegraphics[width=\linewidth]{plots/perf_mem_act_NG.png}
%\end{subfigure}
\caption{The plots show the weak scaling of the 5B (top left), and 15B (top center) model architectures. The 5B does not fit on a single GPU on Frontier, while the 15B needs at least 4 GPUs to fit. Also, the top right and the bottom left plots show the memory usage of the HYBRID and FULL\_SHARD FSDP modes respectively, for the two model architectures. The bottom left plot shows the GPU power, memory, and utilization trace for the different sharding strategies, on 32 node runs, for the 5B model using the \textit{rocm-smi} utility. The average image-per-second (ips) is measured with a local batch-size of 32 for different FSDP sharding strategies. The dash lines (ideal) show the ips drawn from an ideal linear scaling scenario without any additional costs as we scale. The memory usage for HYBRID schemes is constant as we use more nodes, while the for FULL\_SHARD and SHARD\_GRAD it is not.}
\label{fig:weak_scale_NGPUs}
\end{figure*}

Figure \ref{fig:weak_scale_NGPUs} shows the measured image-per-second performance for different FSDP sharding strategies for two models that do not fit on a single GPU: ViT-5B and ViT-15B. The ViT-5B can fit on two GPUs using FSDP's model sharding, while the ViT-15B can fit on four GPUs. We used a local batch size of 32 for both models, which represents a realistic workload for production runs. 

For both models the FULL\_SHARD method performs better as we scale, compared to Figure \ref{fig:weak_scale_1GPU} where the ViT models were able to fit on a single GPU. Interestingly, for the ViT-5B model the HYBRID\_8GPUs and HYBRID\_16GPUs seem to outperform the HYBRID\_2GPUs and HYBRID\_4GPUs strategies, despite the higher communication overhead expected of the former strategies as compared to the latter ones. The HYBRID\_8GPUs and HYBRID\_16GPUs configurations imply on larger numbers of communication calls, with all-gather operations passing through a slower network across nodes. Still, these results indicate that distributing the compute was more beneficial in this case in the trade-off with increased communication overhead.

For the ViT-15B model and its minimum configuration of HYBRID\_4GPUs, we observe for the first time the FULL\_SHARD and the SHARD\_GRAD\_OP modes being competitive with the HYBRID\_SHARD modes. In fact, the SHARD\_GRAD\_OP configuration appears to scale significantly better than all other sharding strategies in this scenario. This further corroborates the observation made for the ViT-5B model that the compute load becomes the driving factor as compared to communication costs, which was in contrast the dominant factor for the smaller ViT models. 

The memory footprint is also shown in Figure \ref{fig:weak_scale_NGPUs} for the two models. Since the SHARD\_GRAD\_OP does not share parameters during computation, but rather only gradients and optimizer states, its memory footprint is much larger than the FULL\_SHARD configuration. However, compared to the FULL\_SHARD, the SHARD\_GRAD\_OP shows a better balance between compute and communication for the ViT-15B experiments using Frontier. It is likely that if we scale the 15B to a much larger number of nodes, the SHARD\_GRAD\_OP would keep better performing than the FULL\_SHARD, but at some point communication would potentially take over again and henceforth the HYBRID\_SHARD modes would perform better than sharding across all ranks.

Finally, for a ViT-5B model the Figure \ref{fig:weak_scale_NGPUs} also presents the GPU power, memory, and utilization trace for the AMD GPUs for the different model sharding strategies. The GPU utilization is approximately 100\%, which likely reflects the fact that we run on synthetic data and thus the interactions with CPU are minimum. Still, as we shown before, we are not IO bound in any of our runs. The SHARD\_GRAD\_OP implied on slightly higher power consumption compared to FULL\_SHARD, which is consistent with the higher throughput that we get between the two (1509 versus 1307 image-per-second). It is noteworthy that the HYBRID\_2GPUs, which is the most performing in terms of throughput (1509 ips), and likely the one with less communication calls compared to the other two, has the smallest power footprint.

\subsection{Performance Observations}
We first observed that as we scale the model size, the IO of the application is not a bottleneck, but rather communication cost is what drives the application performance. The BACKWARD\_PRE and the \textit{limit\_all\_gathers} seem to provide the best performance in terms of parameter prefetching. For models that fit on a single GPU, the best data parallel strategy seems to be the HYBRID\_1GPU, where the cost of synchronization even between the closest GPUs seems to be more expensive than the saves in the compute. It is worth noting that we did not try \textit{torch.compile}, which can offer additional optimizations for overlapping compute and communication costs. For models that can fit on two GPUs, model sharding within the node and data parallel \textit{all\_reduce} across nodes seems to be the best choice. Finally, for models that can fit only on half of the Frontier node, the SHARD\_GRAD\_OP seems to scale better than any other FSDP mode. 

%\section{Model scaling impacts on downstream tasks}
\section{Downstream evaluation as model scales}

In this section, we evaluate the performance of pretrained ViT-Base, Vit-Huge, ViT-1B and ViT-3B models for downstream adaptation on image classification datasets. Specifically, similar to \cite{wang2022rvsa} we first pretrain the models by pairing a MAE formulation with the 1M remote sensing images composing the MillionAID dataset \cite{long2021millionaid}. We then perform linear probing of the four models for scene classification on three independent datasets.

Two main reasons are behind our choice for focusing on linear probing experiments rather than fine-tuning. First, linear probing implies on fewer task specific parameters and training, relying more strongly on the quality of the ideally generalizable features extracted by pretrained FMs. Moreover, fine-tuning results reported in the related literature for these datasets are nearly saturated (above $95\%$ \cite{wang2022rvsa, sun2022ringmo}), such that any performance gains would be rather marginal and could potentially arise due to overfitting of a few wrong labels rather than providing insights on the quality of features being extracted by the pretrained models.

\subsection{Datasets}

Table \ref{table:datasets} summarizes the four datasets used. The MillionAID dataset is one of the largest publicly available geospatial databases, with more than 1M million non-overlapping RS images spanning across 51 classes. Similar to related works \cite{wang2022rvsa}, we used $990K$ images for pretraining and then randomly selected a total of $10K$ images across all categories for downstream adaptation and evaluation.

In addition to MillionAID, we followed \cite{sun2022ringmo} and \cite{wang2022rvsa} and assessed model downstream capabilities across the UC Merced Land Use (UCM) \cite{Yang2010UCM}, the Aerial Image Dataset (AID) \cite{xia2017aid}, and the NWPU-RESISC45 \cite{cheng2017nwpu} collected by the Northwestern Polytechnical University. 
% We also used the domain adaptation dataset, LoveDA [FIX], to evaluate the semantic segmentation downstream task. LoveDA has two larger categories, urban and rural areas, were for this work we focused only on the seven segmentation classes: buildings, roads, water, barren land, forests, agriculture, and background.

\begin{table}[h!]
\caption{The different datasets used for pretraining and linear-probing + evaluation on downstream tasks}
\centering
\begin{tabular}{ |c|c| }
 \hline
   \multicolumn{2}{|c|}{\textit{Pretraining}} \\
   \hline
 \textbf{Datasets} & \textbf{Training Samples} \\
 \hline
 MillionAID   & 990848   \\
 \hline
%   \hline
%  \multicolumn{4}{|c|}{Semantic Segmentation Dataset} \\
%  \hline
%  LoveDA &   2522  & 1669  & 7 \\
%   \hline
\end{tabular}
\vspace{10pt}

\begin{tabular}{ |c|c|c|c| }
 \hline
   \multicolumn{4}{|c|}{\textit{Image Classification}} \\
   \hline
 \textbf{Datasets} & \textbf{Training Samples}  & \textbf{Testing Samples} & \textbf{Classes} \\
 \hline
%  \multicolumn{4}{|c|}{\textit{Pretraining}} \\
%  \hline
%  MillionAID   & 990848    &  &   \\
%  \hline
%  \multicolumn{4}{|c|}{\textit{Image Classification}} \\
%  \hline
 MillionAID &   1000  & 9000  & 51 \\
 UCM &   1050  & 1050  & 21 \\
 AID &   2000  & 8000  & 30 \\
 NWPU &   3150  & 28350  & 45 \\
 \hline
%   \hline
%  \multicolumn{4}{|c|}{Semantic Segmentation Dataset} \\
%  \hline
%  LoveDA &   2522  & 1669  & 7 \\
%   \hline
\end{tabular}
\label{table:datasets}
\end{table}

\subsection{Pretraining}

\begin{figure}[h!]
\centering
  \includegraphics[width=0.85\linewidth]{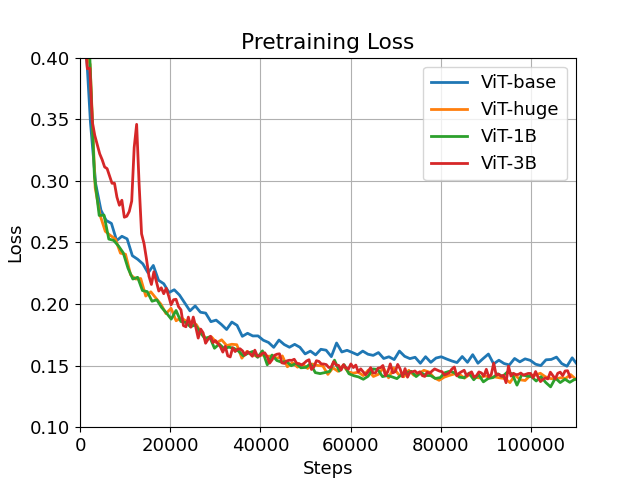}  
\caption{The MAE pretraining loss of the four ViT models, base, huge, 1B, and 3B, as a function of steps, for an approximate total of 100 epochs.}
\label{fig:pre-training-loss}
\end{figure}

In the following paragraphs, for shortness we adopt the prefix ViT to refer to our ViT models pretrained by means of MAE using MillionAID images (e.g., ViT-Base refers to a ViT-Base model pretrained through MAE). Our code is based on the original MAE work \cite{he2022masked}. Figure \ref{fig:pre-training-loss} shows the pretraining loss for all four models, indicating lower training losses for the ViT-Huge, ViT-1B and ViT-3B models as compared to the ViT-Base. The hyper-parameters for all the models were kept identical for a fair comparison between them, with an input image size of $512\times512$ pixels, a base learning-rate of $1.5e-4$, and weight-decay of $0.05$ with the AdamW optimizer. A global batch size of $2048$ was used with the FSDP NO\_SHARD strategy applied, with a local batch size of $32$ and a $75\%$ mask ratio of the original image used for all models. 

%\begin{figure}[h!]
%  \includegraphics[width=\linewidth]{plots/plot_reco_all.png}
%\caption{Masked Autoencoder Visualization}
%\label{fig:pre-training-reco}
%\end{figure}

\subsection{Linear probing and image classification results}

% \begin{table}[h!]
% \centering
% \caption{Linear Probing results across different RS image classification datasets}
% \addtolength{\tabcolsep}{-0.1em}
% \begin{tabular}{ |c|c|c|c|c|c| }
%  \cline{3-6}
%  \multicolumn{2}{c|}{} & \multicolumn{4}{c|}{\textbf{Top1 Acc ($\%$) on}} \\
%  \hline
%  \textbf{Model} & \textbf{\makecell{Pretrain \\ epochs}} & \textbf{\textit{UCM}} & \textbf{\textit{AID}} & \textbf{\textit{NWPU}} & \textbf{\textit{{MillionAID}}} \\
%  \hline
%  MAE-Base\cite{wang2022rvsa}$^\dag$ & 400 & 49.90  & 61.70 & 61.28 & \\
%  MAE-Base & 400 & 45.17 & 52.11 & 54.28 & 47.20 \\
%  MAE-Base & 100 & 40.62 & 41.72 & 42.40 & 41.31 \\
%  MAE-Huge & 100 & 50.00 & 60.78 & 57.24 & 53.28 \\
%  MAE-1B & 100 &  57.10 & 68.89 & 64.35 & 59.14 \\
%  MAE-3B & 100 & 74.05 & 79.96 & 76.43 & 72.98 \\
%  \hline
% \end{tabular}
% \raggedleft
% \footnotesize{$^\dag$results in \cite{wang2022rvsa} are obtained for different data splits as discussed in the text.}\\
% \label{table:results}
% \end{table}

\begin{figure*}[h!]
\centering
\begin{minipage}[t]{1.25\textwidth}
  \centering
  \hspace*{-5cm}\includegraphics[width=1.\linewidth]{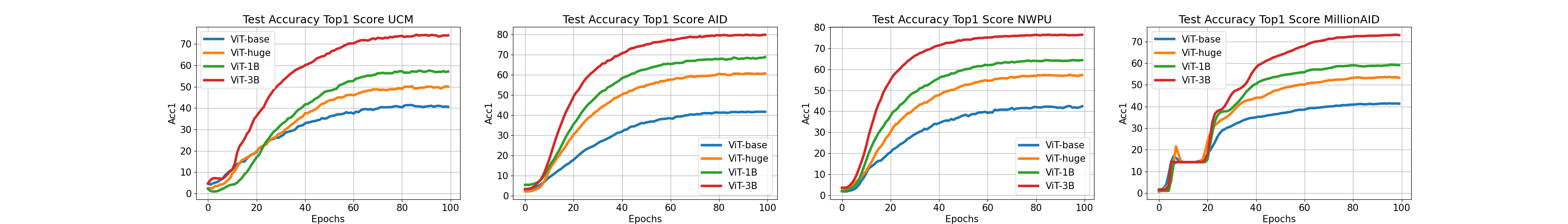}
\end{minipage} \hfill
\begin{minipage}[t]{1.25\textwidth}
  \centering
  \hspace*{-5cm}\includegraphics[width=1.\linewidth]{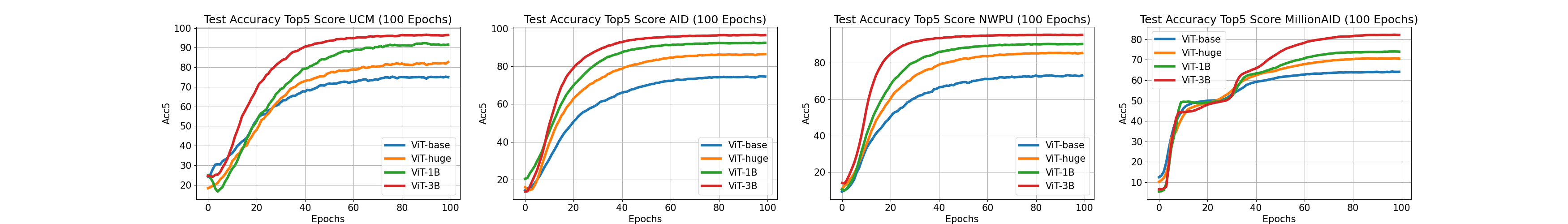}
\end{minipage}\hfill
%\begin{subfigure}{1.25\textwidth}
%  \centering
%  \hspace*{-5cm}\includegraphics[width=1.\linewidth]{plots/linprob_test_loss.png}
%\end{subfigure}\hfill
\caption{The top1 and top5 classification accuracy, of the linear probe classification for the four MAE pretrained ViT models: base, huge, 1B and 3B for all testing datasets. The 100 epochs checkpoint was used from the pretraining, for each model architecture, while keeping all the hyper-parameters constant for the classification task. The training and testing dataset split is as shown in table \ref{table:datasets}}
\label{fig:lin-prob-all}
\end{figure*}

After the self-supervised pretraining, we used linear probing for downstream adaptation of the MAE-pretrained ViT models for image classification. Following common practice \cite{he2022masked}, the LARS \cite{you2017large} optimizer was used, with a base learning rate of 0.1 and no weight decay. The MLP heads of the model were replaced by a linear classifier for supervised training while keeping the weights for the rest of the model frozen. The hyper-parameters were kept identical for all models and datasets. A total number of 100 epochs was used for all the measurements. The global batch-size for UCM, AID and NWPU was kept identical with \cite{wang2022rvsa} at 256, while for the MillionAID dataset since there was no reference to compare, we chose 1024 for faster throughput.

We opt for pretraining our models for 100 epochs and fine-tuning for 100 epochs, in contrast to the 400 epochs of pretraining and 200 epochs of fine-tuning adopted for the ViT-Base models reported in \cite{wang2022rvsa}. Rather than pursuing an indiscriminate use of computing resources and energy, we restrict the pretraining of our models as the results obtained already support the hypothesis that significant improvements are obtained as the models are scaled in size. Table \ref{table:results} shows the top1 classification accuracy for the ViT-Base, ViT-Huge, ViT-1B and ViT-3B for all four image classification datasets.

\begin{table}[h!]
\centering
\caption{Linear Probing results across different RS image classification datasets}
\addtolength{\tabcolsep}{-0.4em}
\begin{tabular}{ |c|c|c|c|c|c| }
 \cline{3-6}
 \multicolumn{2}{c|}{} & \multicolumn{4}{c|}{\textbf{Top1 Acc ($\%$)}} \\
 \hline
 \textbf{Model} & \textbf{\makecell{Pretrain \\ epochs}} & \textbf{\textit{\makecell{UCM\\(TR=$55\%$)}}} & \textbf{\textit{\makecell{AID\\(TR=$28\%$)}}} & \textbf{\textit{\makecell{NWPU\\(TR=$19\%$)}}} & \textbf{\textit{{MillionAID}}} \\
 \hline
 \makecell{ViT-Base\\\cite{wang2022rvsa}}& 400 & 49.90  & 61.70 & 61.28 & \\
 \hline
\end{tabular}
\\
\vspace{10pt}
% \addtolength{\tabcolsep}{-0.1em}
\begin{tabular}{ |c|c|c|c|c|c| }
 \cline{3-6}
 \multicolumn{2}{c|}{} & \multicolumn{4}{c|}{\textbf{Top1 Acc ($\%$)}} \\
 \hline
 \textbf{Model} & \textbf{\makecell{Pretrain \\ epochs}} & \textbf{\textit{\makecell{UCM\\(TR=$50\%$)}}} & \textbf{\textit{\makecell{AID\\(TR=$20\%$)}}} & \textbf{\textit{\makecell{NWPU\\(TR=$10\%$)}}} & \textbf{\textit{{MillionAID}}} \\
 \hline
 ViT-Base & 400 & 45.17 & 52.11 & 54.28 & 47.20 \\
 ViT-Base & 100 & 40.62 & 41.72 & 42.40 & 41.31 \\
 ViT-Huge & 100 & 50.00 & 60.78 & 57.24 & 53.28 \\
 ViT-1B & 100 &  57.10 & 68.89 & 64.35 & 59.14 \\
 ViT-3B & 100 & 74.05 & 79.96 & 76.43 & 72.98 \\
 \hline
\end{tabular}
\label{table:results}
\end{table}

% We kept the original split that was given from the database we download the data from, and as far as we can tell there is no an established split for those three datasets.

Importantly, results in Table \ref{table:results} clearly reveal improvements in image classification performance as we scale the model size, for all four datasets evaluated. Remarkably, we observe more than $30\%$ improvement across all four datasets when going from a 100M parameter to a 3B parameter model. 

The differences in performance we observe between the ViT-Base reported in \cite{wang2022rvsa} and ours are most likely due to differences in data splits used for each dataset, as we adopt more rigorous training/test splits (i.e., fewer training samples). We denote TR the ratio of training samples used for each dataset: while we adopt $TR=50\%$ for UCM, $TR=20\%$ for AID and $TR=10\%$ for NWPU, results in \cite{wang2022rvsa} were obtained with $TR=55\%$ for UCM, $TR=17\%$ for AID and $TR=19\%$ for NWPU. Still, the pattern of performance improvement as our models are scaled in size is evident, with results obtained by our ViT-Huge and larger models surpassing these numbers despite relying on fewer training samples and pretraining epochs.

Figure \ref{fig:lin-prob-all} expands the Table \ref{table:results} by displaying the classification accuracy as a function of the number of linear probing training epochs, for the top1 and top5 classification accuracies. These curves reveal that top1 accuracy improvements become evident even as early as after 10 training epochs for UCM, AID and NWPU datasets, while the top5 accuracy follows a slower trend of improvement. In contrast, it is interesting to note how improvements for the MillionAID dataset occur rather later during training. The constant accuracy between 5 to 20 epochs, could possibly been improved by choosing a more aggressive learning rate decay, compared to the other datasets. Since MillionAID samples were also used for pretraining (but splitting between training and testing), the linear probing training samples in this case come from the same data-distribution as the pretraining samples. We conjecture this is the main factor explaining this difference with respect to the behavior observed for the other three datasets.

% while we only use 10\% of the labels for the linear-probing stage, we don't see any improvement below 10 epochs.

\section{Conclusion}
Foundation models have emerged with desirable properties to integrate and synthesize vast amounts of knowledge toward new AI generalization capabilities. However, these models are yet to be fully explored and deployed in geospatial applications. This lag is due to multiple reasons: (i) these large models require vast HPC training resources that are not afforded to many in the community, (ii) the barrier of entry to training on HPC leadership facilities is high as the current hardware is very specialized and the expertise on effectively leveraging them remains restricted to few organizations, and (iii) pretraining billion-scale models on large RS data requires optimized strategies to data workflows. This study sought to address all three of these challenges and provide a practical guide for training billion parameter size ViT models on HPC systems, using only native PyTorch's distributed library. We provide baselines for various size ViT models, and discuss compute and communication costs to consider in training FMs for the largest geospatial application workloads to date. We also investigate the bottlenecks on distributing ViT training on the Frontier HPC system while scaling to various size models. Lastly, we evaluate our models via linear-probing and show gains up to $+30\%$ on remote sensing imagery classification tasks across three independent datasets. As much as we share new insights on training billion-scale FMs for RS, we also identify new research avenues that are worth following up in future research. Envisioned next steps include evaluation of model capabilities across additional downstream tasks (e.g., object detection and semantic segmentation), and under configurations such as few-shot learning to unveil potential properties emerging as we scale our models into multiple billions of parameters.

\section*{Acknowledgments}
A.T. would like to thank Less Wright from PyTorch team for the valuable discussions. This manuscript has been authored by UT-Battelle, LLC, under contract DE-AC05-00OR22725 with the US Department of Energy (DOE). The US government retains and the publisher, by accepting the article for publication, acknowledges that the US government retains a nonexclusive, paid-up, irrevocable, worldwide license to publish or reproduce the published form of this manuscript, or allow others to do so, for US government purposes. DOE will provide public access to these results of federally sponsored research in accordance with the DOE Public Access Plan (http://energy.gov/downloads/doe-public-access-plan). This research used resources of the Oak Ridge Leadership Computing Facility, which is a DOE Office of Science User Facility supported under Contract DE-AC05-00OR22725.

\bibliographystyle{unsrt} 

\bibliography{main}

\end{document}